# Electric field measurements made on a robotic platform

Karen L. Aplin[1] and Zihao Xiong[2]
[1]Dept. of Aerospace Engineering
University of Bristol, UK
e-mail: karen.aplin@bristol.ac.uk

[2]Bristol Robotics Laboratory, *now at* University of Nottingham, Ningbo

*Abstract*—This presentation reports the first known data from a field mill mounted on a ground-based robotic platform. The robot's motor and electrostatic charging of its wheels do not perturb the field mill data, and electric field varies smoothly whilst the robot is moving. Test measurents under a charged polystyrene plate are reduced in variability by a factor of 2 compared to a hand-held field mill. This technology has potential for autonomous measurements in inaccessible or hazardous environments.

## I. Introduction

Electric field is a fundamental parameter in electrostatics, indicating, for example, the presence of lightning, charged clouds or dust in the atmosphere, or other active charging processes, whether natural or industrial. The vast majority of electric field measurements made so far have been from a static platform, with some mobile measurements from airborne platforms, largely for atmospheric research. These can be uncontrolled, such as a rocket or free balloon, or controlled, such as a plane [1]. Recent technological improvements have led to the increased use of autonomous mobile platforms such as UAVs for environmental measurements [2], and increasingly for atmospheric electricity, both for charge injection for cloud modification [3], and electric field sensing [4].

Ground-based robotic electrostatic measurements remain unexplored, but could have a range of applications, for example, using the electric field distortions from buildings and other objects as a "sixth sense" for navigation [5], or for providing more information for atmospheric electricity research [6]. In these applications the fields are sufficiently small that stray charging effects may be important, and sampling of a broad spatial region is required to make a good-quality measurement. Ground-based mobile electric field measurements are also being studied for future Mars rovers and robotic swarms. Here we mount a field mill on a ground-based robot and describe some simple laboratory experiments to verify its performance.

## II. Methodology

In this proof of principle experiment, a JCI140 electric field mill (FM) was mounted on a Pioneer 2-DX robot, Figure 1. FM technology was chosen as a mechanically robust and readily available solution able to respond to a wide range of electric fields. The Pioneer 2-DX robot was also a readily available and well-established technology, programmed, tested and controlled using the Gazebo simulator and the Robot Operating System (ROS).

The experiments reported here were carried out under a test environment provided by a charged 1 x 1m polystyrene sheet. The sheet was initially charged by rubbing with a glass or rubber rod a fixed number of times, and each comparison run was carried out rapidly, within a few minutes, to ensure the charge on the sheet remained approximately constant during the measurements. The sheet was then charged again using the standard procedure, and the experiment repeated multiple times.

### A. Location of field mill on robot

An important preliminary experiment was to establish the best location for the FM on the robot. This was tested by mounting the FM to four different positions on the chassis (Fig 1), under the centre of the charged plate, and comparing the measurements of potential to those obtained with a FM held at arms' length at the same relative position.

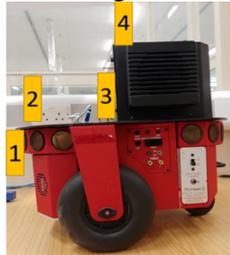

Fig. 1. Pioneer 2-DX robot showing four trial locations for the field mill in yellow.

To determine whether the robot's motor affected the FM data via electromagnetic interference or vibration, the experiments were repeated with the robot switched both on and off. Effects of tribocharging the robot's rubber wheels, which could occur during motion, were also investigated by deliberately charging the wheels with a rubber or glass rod. The presence of wheel charge was verified with another FM, and the wheels were grounded to discharge them before the next run. The results are shown in Fig 2.

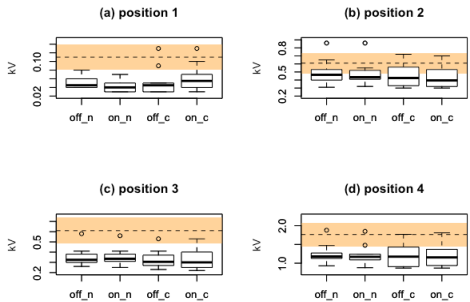

Fig. 2. Potential measured with the field mill at different locations on the robot, shown in Fig 1 (a) position 1, aperture flush with front (b) position 2, front (c) position 3, back (d) position 4, top. The potential under the centre of the polystyrene plate was measured with the robot switched off, and then on, with its wheels in their natural charge state after grounding (_n) and repeated with its wheels charged by rubbing (_c). The dotted line denotes the median of the handheld FM measurements and the shading two standard errors either side.

Figure 2 shows that the power state of the robot's motor barely affected the measurements, and the charge of its wheels only affected the data in position 1 (fig 2a), when the FM was located closest to the wheels. This verifies that, with appropriate positioning, the FM can be operated on the moving robot without its measurements being perturbed. Position 4 (fig 2d) was the most sensitive, probably due to its enhancement of local field lines. In terms of the best position for the FM on the robot, the potential measured in position 2 (fig 2b) was closest to the hand-held FM results, whilst showing minimal effects of power state and wheel charge.

### III. TESTS WITH MOVING ROBOT

In this experiment the robot, with the FM located at position 2 on the robot as defined above, was moved to five positions under the centre of the charged plate (layout shown in Fig 3) and the potential measured at each point. For comparison the experiment was repeated with manual measurements with the FM located at arm's length at the same position and height under the plate.

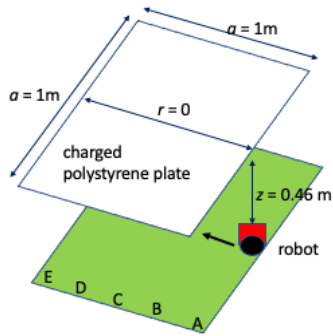

Fig. 3. Moving robot experiment setup. The robot was moved under the centre line of a charged polystyrene plate of side $a$ = 1m, where $r$ = 0 is defined as the centre of the plate. In practice, the potential on the field mill was measured at positions A-E which were labelled on the edge of the polystyrene sheet.

The measurements were compared with a theoretical expression for the potential $V$ [7] with distance $r$ along the centre of a square plate of side $a$, at a fixed perpendicular distance $z$ away of 0.462m, where the plate has uniform surface charge density $\sigma$:

$$V(r) = \frac{z\sigma}{\pi\epsilon_0} \tan^{-1}\left(\frac{a^2}{4r\sqrt{a^2+r^2}}\right) \quad (1)$$

The measurements were fitted to equation (1) using a non-linear least squares fit, weighted by the standard error, with the data and fit shown in Fig. 4.

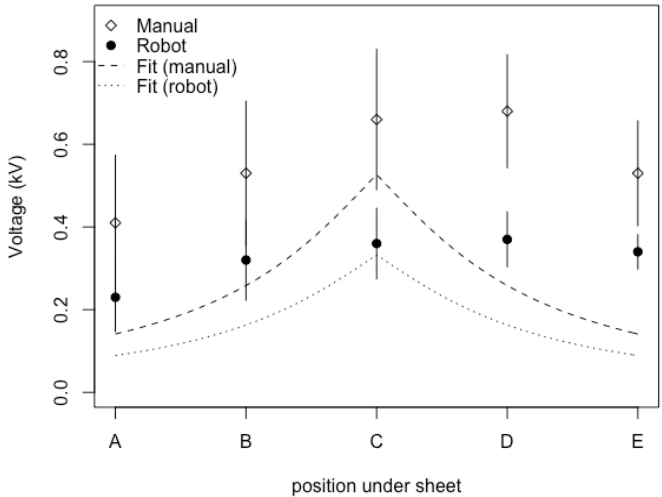

Fig. 4. Potential measured with the FM robot-mounted and handheld at each position over five experimental runs, with the standard error of the mean shown as error bars on the data points. The dashed lines represent non-linear least squares fit to each dataset.

The FM signal changed smoothly and uniformly as the robot was moving. The robot-mounted measurements show less variability than the manual data, with a fractional standard error of 20-40% for the hand-held FM compared to 12-36% for the robot. This suggests that the robot provides a more repeatable way to make electrostatic measurements than a hand-held FM. The theoretical relationship is approximately met for both hand-held and robotic data, with slightly better fitting statistics for the robot, but the results are non-ideal, particularly at the edges. This is likely due to the small size of the polystyrene sheet increasing edge distortion effects, and both spatial and temporal variations in surface charge on the sheet, which is assumed to be uniform in eq (1). The fit between $V$ and $r$ from equation (1) can be used to estimate the surface charge on the sheet from both handheld and robotic measurements and the results are consistent to within the errors in the fit, ~10 pCm$^{-2}$.

IV. CONCLUSION

This paper demonstrates that a field mill can be successfully deployed on a moving robot without its motion or wheel charging perturbing the measurements. A preliminary test under a charged polystyrene sheet showed that variability can be improved by a factor of 2 compared to hand-held measurements. More consistent results could be readily obtained by using a smaller FM [8] and robot such as the Pololu chassis with the Romi 32U4 control board. In these circumstances, when the FM and the robot are closer in mass than the tests reported here (robot mass of 9 kg compared to 200 g for the FM) care may need to be taken to reduce vibration effects such as use of passive isolation mounts. Variability in the fields to be measured could be reduced by applying a fixed voltage to a larger metallic plate. The position of the FM on the robot could also be optimized with electrostatic modelling.

Electrostatics is notorious for its measurement variability, so enhancing repeatability by using a programmable moving platform could yield significant improvements. The reduction in variability provided by robotic measurements could be relevant for fair weather atmospheric electricity measurements, and other environments with relatively low-magnitude but highly variable fields. We believe this is the first time a field mill has been deployed on a robot; this technique has particular potential for autonomous measurements in remote or hazardous environments such as on other planets or within industrial facilities.

## REFERENCES

[1] K.A. Nicoll, "Measurements of atmospheric electricity aloft", *Surv. Geophys*. 33, 5, 991-1057, 2012
[2] E.J. Liu, A. Aiuppa, A. Alan, et al., "Aerial strategies advance volcanic gas measurements at inaccessible, strongly degassing volcanoes". *Science Advances*, 6 (44), eabb9103, 2020
[3] R.G. Harrison, K.A. Nicoll, D.J. Tilley, et al., "Demonstration of a remotely piloted atmospheric measurement and charge release platform for geoengineering", *J. Atmos. Ocean. Tech*., 38, 1, 63-75, 2021
[4] Y. Zhang, E. Duff, A. Agundes, et al., "Small UAV Airborne Electric Field Measurements", *Proc. 24th International Lightning Detection Conf. and 6th International Lightning Meteorology Conf*., San Diego, CA, USA, 18-21 April 2016
[5] Y. Li, W. Zhang, P. Li, et al., "A method for autonomous navigation and positioning of UAV based on electric field array detection," *Sensors*, 21, 4, 1146, 2021
[6] R.G. Harrison, K.A. Nicoll and K.L. Aplin, "Evaluating stratiform cloud base charge remotely", *Geophys. Res. Letts.,* 44, 12, 6407-6412, 2017
[7] Anon, "Electric field due to a uniformly charged finite rectangular plate" [Online] https://physics.stackexchange.com/questions/176747/electric-field-due-to-a-uniformly-charged-finite-rectangular-plate, accessed 2 June 2020
[8] R.G. Harrison and G.J. Marlton, "Fair weather electric field meter for atmospheric science platforms" *J. Electrostatics* 107, 103489, 2020